\documentclass{article}

   \PassOptionsToPackage{numbers, compress}{natbib}

\usepackage[final]{neurips_2020}

\usepackage{neurips_2020}
\usepackage[utf8]{inputenc} 
\usepackage[T1]{fontenc}    
\usepackage{hyperref}       
\usepackage{url}            
\usepackage{booktabs}       
\usepackage{amsfonts}       
\usepackage{nicefrac}       
\usepackage{microtype}      
\usepackage{mdframed,multirow,hyperref,booktabs}
\usepackage{subfigure,graphicx}
\usepackage{multicol}
\usepackage[all]{xy}
\usepackage{xcolor}
\usepackage{graphicx}
\usepackage[linesnumbered,ruled,vlined]{algorithm2e}
\usepackage{algpseudocode}
\usepackage{amsmath}
\usepackage{float}

\title{Amortized Variational Deep Q Network}

\author{%
  Haotian Zhang, Yuhao Wang, Jianyong Sun\thanks{Corresponding author: Jianyong Sun}, Zongben Xu\\
  School of Mathematics and Statistics\\
    Xi'an Jiaotong University\\
  \texttt{\{zht570795275, wyhwhy\}@stu.xjtu.edu.cn, \{jy.sun, zb.xu\}@xjtu.edu.cn} \\
}

\begin{document}
\maketitle

\begin{abstract}
Efficient exploration is one of the most important issues in deep reinforcement learning. To address this issue, recent methods consider the value function parameters as random variables, and resort variational inference to approximate the posterior of the parameters. In this paper, we propose an amortized variational inference framework to approximate the posterior distribution of the action value function in Deep Q Network. We establish the equivalence between the loss of the new model and the amortized variational inference loss.  We realize the balance of exploration and exploitation by assuming the posterior as Cauchy and Gaussian, respectively in a two-stage training process. We show that the amortized framework can results in significant less learning parameters than existing state-of-the-art method. Experimental results on classical control tasks in OpenAI Gym and chain Markov Decision Process tasks show that the proposed method performs significantly better than state-of-art methods and requires much less training time. 
\end{abstract}

\section{Introduction}

Reinforcement Learning (RL) has achieved great successes in games and robotics control. For example, the agents trained with RL on StarCraft and Go have surpassed the top level of human~\cite{silver2018a,vinyals2017starcraft}.

Deep RL algorithms such as Deep Q Network (DQN)~\cite{mnih2013playing} usually apply dithering exploration strategy, such as $\varepsilon$-greedy~\cite{osband2016deep}, noise injection to actions~\cite{lillicrap2016continuous}, and action-level regularization~\cite{williams1992simple}. These exploration strategies are all based on local perturbation of actions, hence are not likely to lead to large scale behavior needed for efficient exploration~\cite{osband2019deep}, which is a must for real-world problem such as auto-driving due to enormously large search space. 

Recent developments inspired by the duality between control and inference~\cite{todorov2008general} have shown improvement on exploration efficiency, such as VIME~\cite{houthooft2016vime}, BBQ Network~\cite{blundell2015weight}, Bootstrapped DQN~\cite{osband2016deep}, NoisyNet~\cite{Fortunato2017Noisy}, and VDQN~\cite{tang2017variational}. Among these works, either parameters of the value/policy function or the parameters that govern the distribution of the value/policy function are considered as random variables. Variational inference technique is applied to approximate the posterior of these parameters. Experimental results suggested variational inference can indeed result in deep exploration.

In this paper, we propose an amortized variational Deep Q network framework, in which the state-action value function, i.e. Q, is considered to be a random variable. The posterior of Q is approximated by variational inference. Our contributions can be summarized as follows. First, the parameters of the auxiliary posterior distribution of Q are the output of a deep neural network as in the variational autoencoder (VAE)~\cite{kingma2014auto}. Second, to encourage exploration, we add entropy bonus to the DQN loss. The equivalence between the loss function of our new model and the amortized variational inference loss is then established. Third, we propose a two-stage (pre-train and fine-tune) training algorithm. The auxiliary posterior is learned from a family of heavy-tailed Cauchy distribution in the pre-train stage, and Gaussian in the fine-tune stage. Finally, experiments on chain MDP and OpenAI Gym tasks verify the proposed method performs significantly better than DQN, VDQN, and NoisyNet.



\section{Background}\label{Background}

\subsection{Markov Decision Process and Reinforcement learning}

\begin{figure}[htbp]
\centerline{\includegraphics[width=0.6\columnwidth]{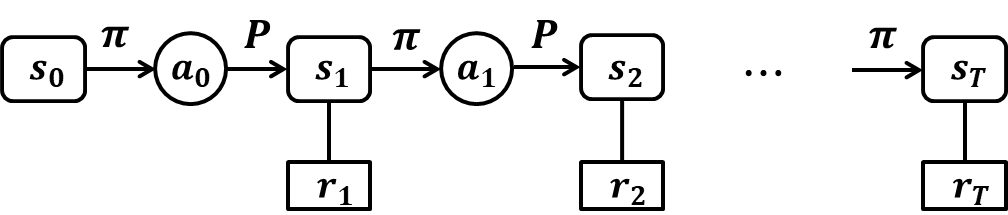}}
\caption{The flowchart of a finite-horizon Markov chain process.}\label{RL_introduction}
\end{figure}

Reinforcement Learning has been playing an important role in the thriving of artificial intelligence. It aims to find a policy for an agent so that it can perform optimally in the environment. RL can be modeled as a Markov Decision Process (MDP). Consider a finite-horizon MDP with state and action space defined by a tuple $(\mathcal{S},\mathcal{A},\mu_0,p,r,\pi,T)$ where $\mathcal{S}\in\mathbb{R}^D$ denotes the state space, $\mathcal{A}\in\mathbb{R}^d$ the action space, $\mu_0$ the initial distribution of the state, $r:\mathcal{S}\rightarrow\mathbb{R}$ the reward, and $T$ the time horizon, respectively. At each time $t$, there are $s_t\in\mathcal{S}$, $a_t\in\mathcal{A}$ and a transition probability $p:\mathcal{S}\times\mathcal{A}\times\mathcal{S}\rightarrow \mathbb{R}$, where $p(s_{t+1}|a_t,s_t)$ denotes the transition probability of $s_{t+1}$ conditionally on $s_t$ and $a_t$. The policy $\pi:\mathcal{S}\times\mathcal{A}\times \{0,1,\cdots\,T\}\rightarrow\mathbb{R}$, where $\pi(a_t|s_t)$ is the probability of choosing action $a_t$ when observing current state $s_t$. Fig.~\ref{RL_introduction} shows the flowchart of the finite-horizon MDP.

At each state $s_t$, the agent takes an action $a_t$ according to present policy $\pi(a_t|s_t)$. Given this action, environment responds with a new state $s_{t+1}$ and an immediate reward $r_{t+1}$. The goal is to find a policy $\pi = p(a_t|s_t)$  so as to maximize the expectation of total rewards $\mathbb{E}(\sum_{t=1}^T \alpha_t r_t) $ where $\alpha_t$ denotes time-step dependent decaying factors. In practice, the decaying factor is set to be the exponential power of a constant, i.e. $\alpha_t=\gamma^t$.

There are many RL algorithms, such as Q-learning, sarsa, deep Q network and policy gradient, which are developed to deal with different environments~\cite{Sutton1998Reinforcement}. Among them, the Q-learning is developed for MDP with discrete state and action space. Its critical idea is to use the action value function $Q(s,a)$ to estimate the reward in case $S_t = s$ and $A_t = a$, by using Bellman equation:
\begin{equation}
Q(s,a) = \mathbb{E}\left(R_{t+1}+ \max_{a_{t+1}}\gamma Q(S_{t+1}, a_{t+1})|S_t = s, A_t = a\right)\label{Bellman equation}
\end{equation}

In practice, Monte Carlo sampling is used to approximate the expectation in Eq.~\ref{Bellman equation}. A number of trajectories $\tau^i\doteq\{s_0^i,r_0^i,a_0^i,\cdots,s_T^i,r_T^i\},\ i=1,\cdots,l$ are sampled and used to update $Q$ as follows:
\begin{equation}
Q(s_t^i,a_t^i)\leftarrow(1-\alpha) Q(s_t^i,a_t^i)+\alpha\left(r_{t+1}^i+\max_{a_{t+1}}\gamma Q(s_{t+1}^i, a_{t+1})\right)
\end{equation}

As $\mathcal{A}$ is discrete and finite, the optimal policy can be regarded as $\pi(a|s)=\arg\max_{a\in\mathcal{A}}Q(s,a)$.

In Deep Q Network (DQN) \cite{mnih2013playing},  the action value function is approximated by a deep neural network with parameter $\theta$ (denoted as $Q_{\theta}(s,a)$). Similar to the Q-learning, the loss function of DQN is
\begin{equation}\label{dqn_loss}
L(\theta)=\frac{1}{l}\frac{1}{T}\sum_{i=1}^l\sum_{t=1}^T\left[Q_{\theta}(s_t^i,a_t^i)-\left(r(s_{t+1}^i)+\max_{a_{t+1}\in {\cal A}}\gamma Q_{\theta}(s_{t+1}^i, a_{t+1})\right)\right]^2
\end{equation}

\subsection{Amortized Variational Inference}

Let $p(x,z)$ be a joint distribution for latent variable $z\in \mathcal{Z}$ and observed variable $x\in\mathcal{X}$. Inference aims to compute the posterior $p(z|x)$. However, the posterior is usually intractable to compute, thus approximation is inevitable. Variational inference uses a family of tractable distributions $\mathcal{Q}$ parameterized by $\psi$ over $z$ and finds a member $q_{\psi_x^*}\in\mathcal{Q}$ by minimizing the Kullback-Leibler (KL) divergence: $q_{\psi_x^*}=\arg\min_{q_{\psi}\in\mathcal{Q}}D_{KL}(q_{\psi}(z)||p(z|x))$ where $q_{\psi}(z)$ is called the auxiliary posterior distribution. For a dataset $\mathcal{D}$, we need to find the auxiliary posterior $q_{\psi_x^*}$ for each latent $z$ associated with each $x\in\mathcal{D}$ under i.i.d. assumpution.

Amortized variational inference (AVI) \cite{gershman2014amortized} assumes that the local variational parameter can be predicted by a parameterized function whose parameters are shared by all data points, Namely, to find a parameterized function $f_{\theta}:\mathcal{X}\rightarrow\mathcal{Q}$ to predict $\psi_x^*$ where $\theta$ is the parameter. Neural network can be readily used as the function $f_{\theta}$ due to its universal function approximation capability~\cite{cybenko1989approximation,hornik1989multilayer}. In literature, such a network is often called inference network. AVI has been applied in VAE~\cite{kingma2014auto}. In VAE, during training, the expectation over $q_{\psi_x}(z)$ is needed for the calculation of the variational lower bound:
\begin{equation}
\log p(x) \geq \mathbb{E}_{q_{\psi_x}(z)} [\log p(x|z)] - D_{KL}(q_{\psi_x}(z)||p(z|x))
\end{equation}
The stochastic back-propagation~\cite{rezende2015variational} is used to approximate the expectation by Monte Carlo gradient estimation, including a reparameterization step and a back-propagation with Monte Carlo step.
\section{Related Work}\label{Related Work}

NoisyNet~\cite{Fortunato2017Noisy} proposes to directly add noises to the parameters of the value/policy function. It is shown that doing so can enable consistent exploration. Bootstrapped DQN~\cite{osband2016deep} applies the bootstrap to approximate the posterior of the value function by applying different heads trained with different bootstrapped data, which entails diverse strategies to encourage exploration, but the improvement is limited and may be compute-intensive when training.

Bayes-by-Backprop Q-Network (BBQN)~\cite{blundell2015weight} randomizes the policy space to achieve the balance between exploration and exploitation. It achieves good performance on dialogue tasks when combined with Replay Buffer Spiking (RBS) trick and pipeline of natural language processing system~\cite{DBLP:conf/aaai/LiptonLG00D18}.

Variational Information Maximizing Exploration (VIME)~\cite{houthooft2016vime} proposes an intrinsic reward based on information bonus to encourage exploration, which is a curiosity-driven exploration strategy. It learns a dynamic model of environment represented by a Bayesian neural network and uses a variational family distribution to approximate posterior and compute bonus.

Variational Deep Q Network (VDQN)~\cite{tang2017variational} shares a similar spirit with BBQN. It proposes a surrogate objective to the Bellman error by adding an entropy term which can explicitly encourages exploration. It is shown that the surrogate objective is equivalent to the variational inference loss. Moreover, the algorithm is interpreted as performing approximate Thompson Sampling. Bayesian neural network is used as the posterior and variational inference subroutines are used to minimize the KL divergence.




Recently, distributional RL methods~\cite{tang2018exploration,moerland2018the,mavrin2019distributional,dabney2017distributional} have been proposed, in which the cumulative reward is considered as random variable and the distributional Bellman equation is applied to iteratively update the distribution to minimize the distance between the target distribution and the predicted distribution.  It is more challenging than computing the expectation of the reward but it can offer more information.
\section{Method}\label{Method}



\subsection{Model}\label{avdqn}

In DQN, $\theta$ in Eq.~\ref{dqn_loss} denotes the parameters of the DNN. In VDQN, $\theta$ is regarded as a random variable. The posterior of $\theta$ of a Bayesian neural network is approximated by variational inference. 

In our work, we consider $Q_{\theta}(a_t^i,s_t^i)$ to be a random variable and borrow the idea of amortized variational inference for the updating of $q(Q_{\theta}(a_t^i,s_t^i))$. Assume $Q_{\theta}(a_t^i,s_t^i) \sim q(Q_{\theta}(a_t^i,s_t^i);\alpha_{\theta}(a_t^i,s_t^i))$ where $\alpha_{\theta}(a_t^i,s_t^i)$ is the parameter of the posterior distribution $q$. The same as in VAE, $\alpha_{\theta}(a_t^i,s_t^i)$ is the output of a deep neural network (DNN) with parameter $\theta$ and input $a_t^i,s_t^i$. Since $Q_{\theta}(a_t^i,s_t^i)$ is a random variable, Eq.~\ref{dqn_loss} can be re-written as follows:
\begin{eqnarray}
\min_{\theta}\sum_{i,t} \mathbb{E}_{Q_{\theta}(a_t^i,s_t^i)\sim q(Q_{\theta}(a_t^i,s_t^i);\alpha_{\theta}(a_t^i,s_t^i))}\left[\left (Q_{\theta}(a_t^i,s_t^i)-\gamma \cdot
\max_{a_{t+1}\in {\cal A}}Q_{\theta}(a_{t+1},s_{t+1}^i)-r(s_{t+1}^i)\right)^2\right]\label{AVDQN_log}
\end{eqnarray}

If we regard $\gamma \max_{a_{t+1}\in {\cal A}}Q_{\theta}(a_{t+1},s_{t+1}^i)+r(s_{t+1}^i)$ as observed data (denoted as $\mathcal{D}_t^i$), and assume the prior $p(Q_{\theta}(a_t^i,s_t^i))\propto1$, Eq.~\ref{AVDQN_log} is equal to
\begin{eqnarray}
\max_{\theta}\sum_{i,t} \mathbb{E}_{Q_{\theta}(a_t^i,s_t^i)\sim q(Q_{\theta}(a_t^i,s_t^i);\alpha_{\theta}(a_t^i,s_t^i))}\left[ \ln P\left(\mathcal{D}_t^i|Q_{\theta}\left(a_t^i,s_t^i\right)\right)\right]\label{log_posterior}
\end{eqnarray}
where $p\left(\mathcal{D}_t^i | Q_{\theta}(a_t^i,s_t^i)\right)=\mathcal{N}\left(\mathcal{D}_t^i|Q_{\theta}(a_t^i,s_t^i),1\right)$ with $Q_{\theta}(a_t^i,s_t^i)$ as the mean and unit variance.

The same as in \cite{tang2017variational}, adding an entropy term over $q(Q_{\theta}(a_t^i,s_t^i);\alpha_{\theta}(a_t^i,s_t^i))$ on Eq.~\ref{log_posterior} to encourage exploration, we obtain the following loss function
\begin{eqnarray}
\max_{\theta}\sum_{i,t} \mathbb{E}_{Q_{\theta}(a_t^i,s_t^i)\sim q(Q_{\theta}(a_t^i,s_t^i);\alpha_{\theta}(a_t^i,s_t^i))}\left[ \ln p\left(Q_{\theta}\left(a_t^i,s_t^i\right)|\mathcal{D}_t^i\right)\right]+\mathcal{H}[q(Q_{\theta}(a_t^i,s_t^i);\alpha_{\theta}(a_t^i,s_t^i))]\label{AVDQN_LOSS}
\end{eqnarray}where $\mathcal{H}[\cdot]$ denotes the entropy. Eq.~\ref{AVDQN_LOSS} can then be rewritten as
\begin{eqnarray}
\min_{\theta}\sum_{i,t} D_{\text{KL}}\left(q(Q_{\theta}(a_t^i,s_t^i);\alpha_{\theta}(a_t^i,s_t^i))||P(Q_{\theta}(a_t^i,s_t^i)|\mathcal{D}_t^i)\right)\label{KL_divergence}
\end{eqnarray}Eq.~\ref{KL_divergence} indicates that Eq.~\ref{AVDQN_LOSS} actually finds $\theta$ to minimize the sum of all the KL-divergences between $q(Q_{\theta}(a_t^i,s_t^i))$ and the posterior $p\left(Q_{\theta}(a_t^i,s_t^i)|\mathcal{D}_t^i\right)$ at the $i$th episode. That is, if we optimize Eq.~\ref{AVDQN_LOSS}, an optimal $\theta^*$ can be obtained to make $q(Q_{\theta^*}(a_t^i,s_t^i);\alpha_{\theta^*}(a_t^i,s_t^i))$ approximate the true posterior.
\begin{figure}[htbp]
\setlength{\belowcaptionskip}{-2pt}
\setlength{\abovecaptionskip}{-4pt}
\centering\includegraphics[width=0.82\columnwidth]{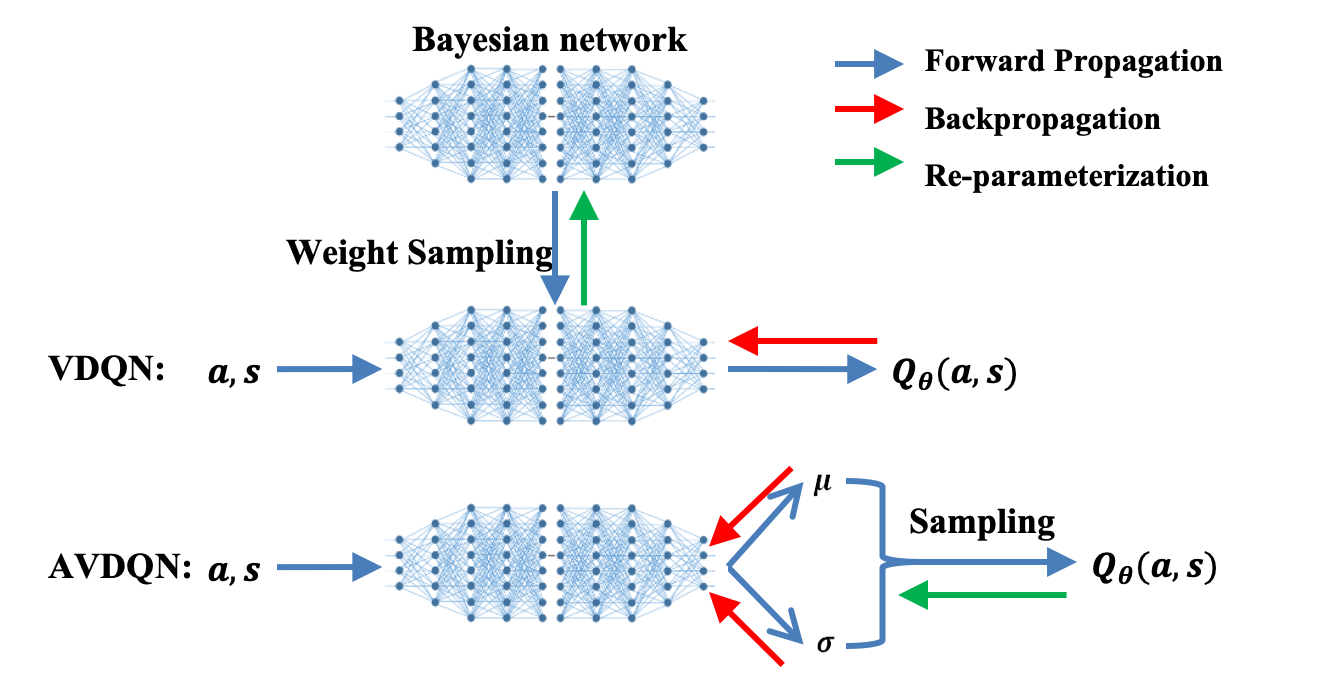}
\caption{Framework of VDQN and AVDQN. Blue (resp. red) arrow indicates the forward propagation (resp. backpropagation) in the neural network pipeline. Green arrow is the reparameterization trick.}\label{framework}
\end{figure}

Fig.~\ref{framework} shows the framework of VDQN and the proposed amortized variational deep Q network (AVDQN). In VDQN, the weights of the Bayesian neural network are random variables. In the forward propagation, a sampled neural network with parameter $\theta'$ is obtained by sampling the weights of the Bayesian network i.i.d. The action-value $Q_{\theta'}(s,a)$ is then obtained by taking $a$ and $s$ as input to the sampled neural network. In AVDQN, the weights $\theta$ of the network are deterministic values, $Q_{\theta}(s,a)$ is considered to be random. The parameters of the posterior of $Q_{\theta}(s,a)$ (in the figure, $\mu$ and $\sigma$ represent the mean and standard deviation, respectively) are the output of the network, while $Q_{\theta}(a,s)$ is the sampled value from ${\cal N}(\mu,\sigma^2)$. In the backpropagation, reparametrization trick is applied differently. In VDQN, it is applied to obtain a set of Bayesian networks, while in AVDQN, it is to obtain different parameters ($\mu$ and $\sigma$) of the auxiliary posterior distribution.

\subsection{Algorithm}\label{tavdqn}

Applying variational inference aims to enhance the exploration ability. However, balancing the exploration and exploitation is also important. Therefore, the training of AVDQN includes a pre-train stage and a fine-tune stage. In the pre-train stage, the posterior $q$ is assumed to be a Cauchy distribution. Cauchy distribution is heavy-tailed which means it is suitable for exploration. In the fine-tune stage, $q$ is considered to be a Gaussian distribution, which is preferable for exploitation.


The training procedure of AVDQN is summarized in Alg.~\ref{alg:tavdqn}. The same as in DQN~\cite{mnih2013playing}, to stabilize training, we use an evaluation network and a target network with parameter $\theta$ and $\theta^-$, respectively. From line~\ref{a6} to \ref{a11}, we sample a $Q_\theta(s_t,a) $ from distribution $q$ for each $a\in {\cal A}$. From line~\ref{a12} to~\ref{a13}, at each time step $t$, we select action by being greedy w.r.t. $Q_\theta(s_t,a)$ and add experience tuple $ \{ s_t,a_t,r_t,s_{t+1} \} $ to buffer $R$. From line~\ref{a14} to \ref{a21}, when updating parameters, we sample a mini-batch of $M$ tuples from the buffer $R$ and use the target network to obtain the target value. From line~\ref{a22} to~\ref{a23}, the gradient of the proposed loss function is computed and used to update parameter $\theta$. In line~\ref{a25}, the target network parameter $\theta^-$ is updated at every $\tau$ steps. In the first $\omega$ episodes, the Cauchy distribution is applied when sampling Q (line~\ref{sampleC1} and~\ref{sampleC2}) where ${\cal C}(\cdot,\cdot)$ denotes the Cauchy distribution. The Gaussian distribution is used in the fine tune stage (line~\ref{a11} and~\ref{sampleN}).
\begin{algorithm}
\caption{Amortized Variational DQN}\label{alg:tavdqn}
\vspace{-1pt}
\LinesNumbered
\KwIn{the target network update period $\tau$; the learning rate $\alpha$; the number of episodes used for pre-train $\omega$; and the discount factor $\gamma$;}
Initialize: parameter $\theta$ and $\theta ^ -$, replay buffer $R \gets \{ \}$, step counter $counter \gets 0$ and $e\gets 0$

\While{not converge} 
{
 $e\gets e+1, t\gets 0$;\\
 \While{\rm{episode not terminated}}
 {
  $counter \gets counter+1$

  \If {$e \le \omega $}
  {
   Compute $\mu_{t}^i,\delta_{t}^i = \text{DNN}(s_t, a_i;\theta)$ for every $a_i \in {\cal A}$; \\\label{a6}
   Sample $Q^i_\theta(s_t,a_i) $ from ${\cal C}\left(\mu_t^i,\delta_t^i\right)$, for every $a_i\in\mathcal{A}$ \label{sampleC1}
  }
  \Else
  {
   Compute $\mu_{t}^i,\sigma_t^i = \text{DNN}(s_t, a_i;\theta)$ for every $a_i \in {\cal A}$; \\
   Sample $Q_\theta(s_t,a_i) $ from ${\cal N}\left(\mu_t,[\sigma_t^i]^2\right)$, for every $a_i\in\mathcal{A}$\label{a11}
   }
  For state $s_t$, set $a_t=\arg\max_{a_i \in {\cal A}} Q^i_\theta(s_t,a_i)$, obtain transition $s_{t+1}$ and reward $r_t$\label{a12}

  Save experience tuple $ \{ s_t,a_t,r_t,s_{t+1} \} $ to buffer $R$\label{a13}

  Sample $M$ tuples $D= \{ s_j,a_j,r_j,s'_j \}$ from $R$\label{a14}

  \If {$e\le\omega$}
  {
    \For{$j = 1\to M$}
  {
   Compute $\mu_{t}^i,\delta^i_{t} = \text{DNN}(s'_j, a_i;\theta^-)$ for all $a_i \in {\cal A}$;\\
   Sample $Q_{\theta^-}(s'_j,a_i)$ from ${\cal C}(\mu^i_{t},\delta^i_{t})$ for $a_i\in\mathcal{A}$\label{sampleC2}
   }
  }
  \Else
  {
  \For{$j = 1\to M$}
  {
   Compute $\mu_{t}^i,{\sigma_t^i} = \text{DNN}(s'_j, a_i; \theta^-)$ for all $a_i \in {\cal A}$; \\
   Sample $Q_{\theta^-}(s'_j,a_i)$ from ${\cal N}\left(\mu_{t}^i,[{\sigma_t^i}]^2\right)$ for all $a_i\in\mathcal{A}$ \label{sampleN}
   }
  }

  Compute the target value for tuples in $D$: $d_j=r_j+\gamma \cdot \max_{a_i}Q_{\theta ^-} (s'_j,a_i), 1\leq j \leq N$\label{a21}

  Take the gradient of Eq.~\ref{AVDQN_LOSS} to obtain $\triangle \theta$\label{a22}

  $\theta = \theta - \alpha \triangle \theta$\label{a23}

  $t \gets t+1$

  \If {$counter$ {\rm mod} $\tau=0$}
  {
   Update the parameters of the target network $\theta ^- \leftarrow \theta$\label{a25}
  }
 }
}

\end{algorithm}
The original DQN samples experiences uniformly from the buffer. In sparse reward scenarios, this may lead to sampling some important but rare experiences with very small probability and results in converging to local optimum~\cite{schaul2015prioritized}. Prioritized Experience Replay (PER)~\cite{schaul2015prioritized} proposes to give each experience a priority. When sampling, experience with higher priority is preferable. The priority is defined either based on proportion or rank. For the proportional priority, the priority of the $i$th experience is $p_i=(\delta_i+\varepsilon)$ where $\delta_i$ is the temporal-difference error. For the rank-based priority, $p_i=\frac{1}{\text{rank}_i}$ where $\text{rank}_i$ is the rank of the transition $i$ when the replay memory is sorted according to $|\delta_i|$. To be more specific, the probability of the $i$th experience to be sampled is $p(i)=\frac{p_i^{\alpha}}{\sum_k {p_k}^\alpha}$. It degenerates to uniform sampling when $\alpha=0$. In our study, the rank-based prioritized replay is used since it is insensitive to outliers and blind to the relative error scales~\cite{schaul2015prioritized} .

\subsection{The Number of Model Parameters}\label{ca}

Assuming the deep neural network has $\ell$ hidden layers, and the $i$th layer has $H_i$ neurons in DQN, VDQN, NoisyNet, and AVDQN. The dimension of the input layer is $I$ and the dimension of the output layer of AVDQN and VDQN is $n_A\cdot n_V$ and $n_V$, respectively, where $n_A$ is the number of parameters of $q(Q)$. In total, the number of parameters of AVDQN is $(I+1)H_1+\sum_{i=1}^{\ell-1}(H_{i}+1)H_{i+1}+n_A(H_{\ell}+1)n_V$. In AVDQN, $n_A=2$ because $q(Q)$ follows Cauchy or Gaussian distribution. In VDQN, each weight has an auxiliary posterior distribution, thus the total number of parameters is $((I+1)H_1+\sum_{i=1}^{\ell-1}(H_{i}+1)H_{i+1}+(H_l+1)n_V)\times n'_A$, where $n'_A=2$ since each weight follows a Gaussian distribution.

In comparison, we see that VDQN has $((I+1)H_1+\sum_{i=1}^{l-1}(H_{i}+1)H_{i+1})\times (n'_A-1)$ more parameters than AVDQN. In case all $H_i$'s are the same ($=H$), VDQN has $(H(\ell-1)+ I + \ell)H$ more parameters than AVDQN.

\section{Experiments}\label{Experiments}


In this section, we test AVDQN on four classic control tasks in OpenAI Gym: CartPole-v0, CartPole-v1, Acrobot-v1, and MountainCar-v0 and four chain MDP tasks with $N=5, 10, 50$ and $100$ (for details about the task please see Appendix). These tasks are very challenging because only through extensive exploration, a proper policy can be learned to control mechanical systems. Particularly, for the CartPole-v0/CartPole-v1 tasks, it is impossible for $\varepsilon$-greedy strategy to learn a policy to balance the pole. On the other hand, learning an optimal policy for chain MDP tasks  becomes much more difficult as $N$ increases~\cite{tang2017variational,osband2016deep}. DQN, VDQN, and NoisyNet are used as baselines in the experiments. 


\subsection{Implementation Details}\label{det}

In all the experiments, we use the fully connected network with two hidden layers and ReLU activation as the inference network. Each hidden layer has 100 units. The size of the mini-batch sampled from the buffer $R$ is $128$. The target network is updated every $100$ time steps. All results are averaged over $5$ different random initializations. The maximum size of the replay buffer is $10^6$. DQN uses $\varepsilon$-greedy strategy where $\varepsilon$ linearly decays from $1$ to $0.01$ as carried out in the original reference. 

On classic control tasks, we use $1500$ episodes to train the model. The learning rate $\alpha$ for DQN, VDQN, NoisyNet, and AVDQN is $10^{-2},10^{-3},10^{-3},10^{-3}$, respectively. The discount factor $\gamma=0.99$ for all the compared algorithms. For the chain MDP tasks, we use $1000$ episodes to train $5,10$ and $3000$ episodes for $N=50,100$. For AVDQN, the last $200$ episodes are used for fine-tune. The learning rate is set constant in the pre-train stage, and set as $\alpha/(1+0.9 \times (e-\omega))$ in the fine-tune stage where $e$ is the training episode, and $\omega$ is the number of episodes used for the pre-train. The discount factor is $\gamma=1$. For NoisyNet, we use independent Gaussian noise. In AVDQN, the rank-based prioritized replay is applied~\cite{schaul2015prioritized}. We save the experience to a priority queue implemented with an array-based binary heap. The heap array was infrequently sorted every $1000$ time steps to prevent the heap becoming too unbalanced.

\subsection{Results}\label{result}

\begin{figure}[htbp]
\subfigure[CartPole-v0]{\includegraphics[scale = 0.4]{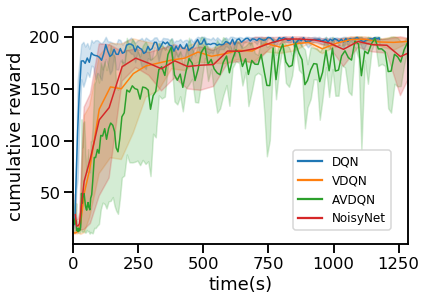}}\qquad%
\subfigure[CartPole-v1]{\includegraphics[scale=0.4]{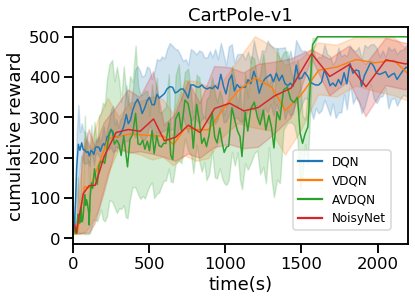}}\\
\subfigure[Acrobot-v1]{\includegraphics[scale = 0.4]{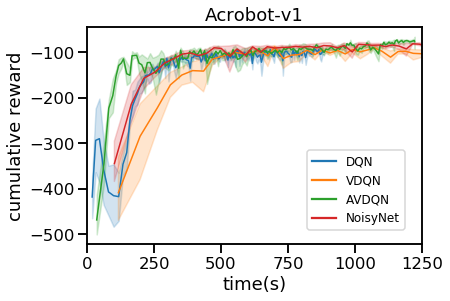}}
\subfigure[MountainCar-v0]{\includegraphics[scale = 0.4]{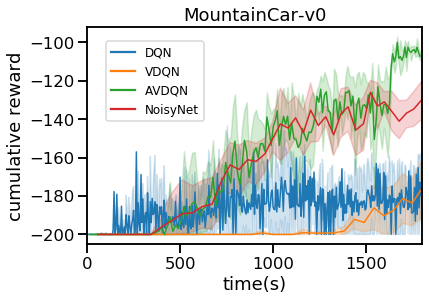}}
\caption{The training curves of DQN, VDQN, NoisyNet, and AVDQN on the four classic tasks within the same running time (in seconds).}\label{fig4}
\end{figure}

\begin{figure}
\subfigure[$N=5$]{\includegraphics[scale=0.4]{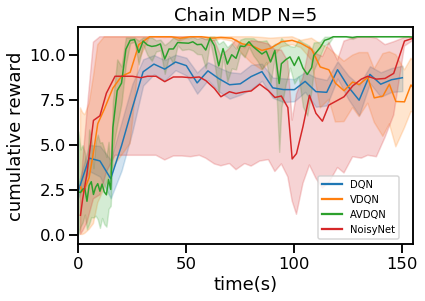}}\qquad
\subfigure[$N=10$]{\includegraphics[scale=0.4]{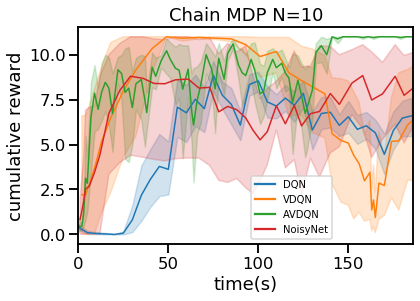}}\\
\subfigure[$N=50$]{\includegraphics[scale=0.4]{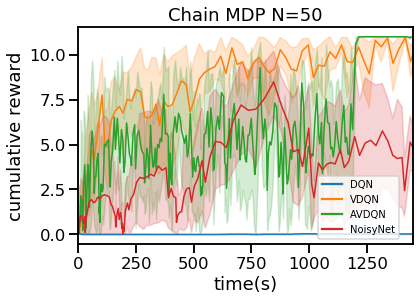}}\qquad
\subfigure[$N=100$]{\includegraphics[scale=0.4]{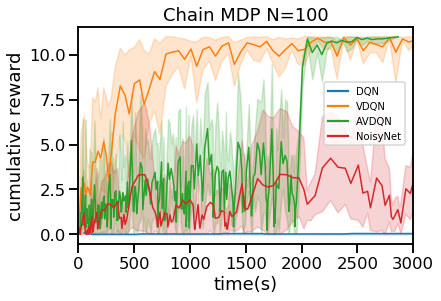}}
\caption{The training curves of DQN, VDQN, NoisyNet, and AVDQN on the four chain MDP tasks averaged over 5 random initializations with same running time.}\label{fig5}
\end{figure}

First we compare the performances of DQN, VDQN, NoisyNet, and AVDQN based on the same running time. Figs.~\ref{fig4} and~\ref{fig5} show the training curves of these compared algorithms. We can see that, AVDQN can reach the highest final reward on all the tasks. Particularly, we find in the chain MDP tasks with $N=5$ and $10$, VDQN is unsteady: there is an obvious declining from $50$ seconds upwards. Possible reason is that VDQN over-emphasizes on exploration so that VDQN cannot converge on some simple tasks. In comparison, on difficult tasks (chain MDP with $N=50$ and $100$) DQN cannot explore deeply while VDQN performs well, but AVDQN performs the best. In addition, NoisyNet reveals unsteady exploration in chain MDP problem when $N$ is large. We may thus conclude that AVDQN can successfully balance the exploration and exploitation.

Table~\ref{tab1} shows the average running time of the compared algorithms within the same number of episodes. It can be observed that AVDQN requires much less time compared with VDQN and NoisyNet. Specifically, training AVDQN is at least two times and at most eight times faster than training VDQN and NoisyNet. Table~\ref{para} shows the number of parameters of DQN, VDQN, NoisyNet, and AVDQN. It is seen that the number of parameters of DQN and AVDQN are similar, and are only half to that of VDQN and NoisyNet. Table~\ref{tab2} shows the final rewards obtained by the compared algorithms within the same running time. From the table we can see that in all the chain MDP tasks, AVDQN has reached the global optimal reward (11) and on the classical control tasks in OpenAI Gym, AVDQN has achieved higher rewards than DQN, VDQN and NoisyNet.
\begin{table}[htbp]
\centering
\caption{The average running time of DQN, VDQN, NoisyNet, and AVDQN on the considered tasks with the same number of episodes (in seconds)}.
    \label{tab1}
    \begin{tabular}{lllllll}
    \toprule
        Tasks & Episodes & DQN & VDQN & NoisyNet & AVDQN \\
    \midrule
        CartPole-v0 & 1500 & 1176 & 7531 & 7955 & 1286 \\
        CartPole-v1 & 1500 & 2350 & 16461 & 16871 & 2196 \\
        Acrobot-v1 & 1500 & 884 & 4241 & 3882 & 1223 \\
        MountainCar-v0 & 1500 & 1241 & 6610 & 5428 & 1790 \\
        MDP $N=5$ & 1000 & 55 & 379 & 389 & 155 \\
        MDP $N=10$ & 1000 & 77 & 544 & 555 & 188 \\
        MDP $N=50$ & 3000 & 796 & 6722 & 7070 & 1448 \\
        MDP $N=100$ & 3000 & 1584 & 16383 & 17193 & 2866 \\
    \bottomrule
    \end{tabular}
\end{table}

\begin{table}[htbp]
\centering
\caption{The number of parameters of DQN, AVDQN, VDQN, and NoisyNet on the considered tasks.}\label{para}
    \begin{tabular}{llllll}
    \toprule
        Tasks & DQN & AVDQN & VDQN & NoisyNet \\
    \midrule
        CartPole-v0 & 10802 & 11004 & 21604 & 21604 \\
        CartPole-v1 & 10802 & 11004 & 21604 & 21604 \\
        Acrobot-v1 & 11103 & 11406 & 22206 & 22206 \\
        MountainCar-v0 & 10703 & 11006 & 21406 & 21406 \\
        MDP $N=5$ & 10902 & 11104 & 21804 & 21804 \\
        MDP $N=10$ & 11402 & 11604 & 22804 & 22804 \\
        MDP $N=50$ & 15402 & 15604 & 30804 & 30804 \\
        MDP $N=100$ & 20402 & 20604 & 40804 & 40804 \\
    \bottomrule
    \end{tabular}
\end{table}

\begin{table}[H]
    \centering
   \caption{The average final rewards in the last 10 episodes of DQN, VDQN, NoisyNet, and AVDQN on the considered tasks under the same running time. }
    \label{tab2}
    \begin{tabular}{llllll}
    \toprule
        Tasks & Time(s) & DQN & VDQN & NoisyNet & AVDQN \\
    \midrule
        CartPole-v0 & 1300 & 198.89 & 196.58 & 185.64 & \textbf{200} \\
        CartPole-v1 & 2350 & 416.71 & 480.33 & 362.64 & \textbf{500} \\
        Acrobot-v1 & 1250 & -104.69 & -104.76 & -92.18 & \textbf{-74.51} \\
        MountainCar-v0 & 1800 & -178.27 & -176.84 & -119.58 & \textbf{-107.09} \\
        MDP $N=5$ & 155 & 9 & 8.62 & 10 & \textbf{11} \\
        MDP $N=10$ & 190 & 8.69 & 6.64 & 8.56 & \textbf{11} \\
        MDP $N=50$ & 1450 & 0.36 & 9.73 & 1.89 & \textbf{11} \\
        MDP $N=100$ & 3000 & 0.04 & 10.67 & 0.24 & \textbf{11} \\
    \bottomrule
    \end{tabular}
\end{table}

\section{Conclusion}\label{Conclusion}

We proposed a framework called Amortized Variational Deep Q Network (AVDQN) to tackle the deep exploration problem in reinforcement learning. In AVDQN, the action value function is considered to be random variables. Under amortized variational inference, an inference network is used to output the parameters of the posterior distribution of $Q$. To train the model, we used the heavy-tailed Cauchy distribution in the pre-train stage to explore the parameter space, and used the Gaussian distribution in the fine-tune stage to exploit for the optimal parameters of the inference network. The experiments on classic control tasks and chain MDP tasks showed that AVDQN performs significantly better than the state-of-the-art RL methods, including DQN, VDQN, and NoisyNet, in terms of the final reward and the training time. 
\section*{Acknowledgements}
This work was partly supported by the National Natural Science Foundation of China (grant no.
11991023, 62076197), the Major Project of National Science Foundation of China (grant no. U1811461), and Key
Project of National Science Foundation of China (grant no. 11690011).
\bibliographystyle{unsrt}
\bibliography{reference}

\begin{thebibliography}{10}

\bibitem{silver2018a}
David Silver, Thomas Hubert, Julian Schrittwieser, Ioannis Antonoglou, Matthew
  Lai, Arthur Guez, Marc Lanctot, Laurent Sifre, Dharshan Kumaran, Thore
  Graepel, et~al.
\newblock A general reinforcement learning algorithm that masters chess, shogi,
  and go through self-play.
\newblock {\em Science}, 362(6419):1140--1144, 2018.

\bibitem{vinyals2017starcraft}
Oriol Vinyals, Timo Ewalds, Sergey Bartunov, Petko Georgiev, Alexander
  Vezhnevets, Michelle Yeo, Alireza Makhzani, Heinrich Kuttler, John Agapiou,
  Julian Schrittwieser, et~al.
\newblock Starcraft ii: A new challenge for reinforcement learning.
\newblock {\em arXiv: Learning}, 2017.

\bibitem{mnih2013playing}
Volodymyr Mnih, Koray Kavukcuoglu, David Silver, Alex Graves, Ioannis
  Antonoglou, Daan Wierstra, and Martin Riedmiller.
\newblock Playing atari with deep reinforcement learning.
\newblock In {\em NeurIPS workshop in Deep Learning}, 2013.

\bibitem{osband2016deep}
Ian Osband, Charles Blundell, Alexander Pritzel, and Benjamin Van~Roy.
\newblock Deep exploration via bootstrapped dqn.
\newblock In {\em NeurIPS}, pages 4033--4041, 2016.

\bibitem{lillicrap2016continuous}
Timothy Lillicrap, Jonathan~J Hunt, Alexander Pritzel, Nicolas Heess, Tom Erez,
  Yuval Tassa, David Silver, and Daan Wierstra.
\newblock Continuous control with deep reinforcement learning.
\newblock In {\em ICLR}, 2016.

\bibitem{williams1992simple}
Ronald~J Williams.
\newblock Simple statistical gradient-following algorithms for connectionist
  reinforcement learning.
\newblock {\em Machine Learning}, 8(3):229--256, 1992.

\bibitem{osband2019deep}
Ian Osband, Benjamin Van~Roy, Daniel Russo, and Zheng Wen.
\newblock Deep exploration via randomized value functions.
\newblock {\em Journal of Machine Learning Research}, 20(124):1--62, 2019.

\bibitem{todorov2008general}
Emanuel Todorov.
\newblock General duality between optimal control and estimation.
\newblock In {\em 47th IEEE Conference on Decision and Control}, pages
  4286--4292, 2008.

\bibitem{houthooft2016vime}
Rein Houthooft, Xi~Chen, Yan Duan, John Schulman, Filip De~Turck, and Pieter
  Abbeel.
\newblock Vime: Variational information maximizing exploration.
\newblock In {\em NeurIPS}, pages 1109--1117, 2016.

\bibitem{blundell2015weight}
Charles Blundell, Julien Cornebise, Koray Kavukcuoglu, and Daan Wierstra.
\newblock Weight uncertainty in neural networks.
\newblock {\em arXiv: Machine Learning}, 2015.

\bibitem{Fortunato2017Noisy}
Meire Fortunato, Mohammad~Gheshlaghi Azar, Bilal Piot, Jacob Menick, Ian
  Osband, Alex Graves, Vlad Mnih, Remi Munos, Demis Hassabis, and Olivier
  Pietquin.
\newblock Noisy networks for exploration.
\newblock In {\em ICLR}, 2018.

\bibitem{tang2017variational}
Yunhao Tang and Alp Kucukelbir.
\newblock Variational deep q network.
\newblock In {\em NeurIPS workshop on Bayesian Deep Learning}, 2017.

\bibitem{kingma2014auto}
Diederik~P Kingma and Max Welling.
\newblock Auto-encoding variational bayes.
\newblock In {\em ICLR}, 2014.

\bibitem{Sutton1998Reinforcement}
R~Sutton and A~Barto.
\newblock {\em Reinforcement Learning:An Introduction}.
\newblock MIT Press, 1998.

\bibitem{gershman2014amortized}
Samuel~J Gershman and Noah~D Goodman.
\newblock Amortized inference in probabilistic reasoning.
\newblock {\em Cognitive Science}, 36(36), 2014.

\bibitem{cybenko1989approximation}
George Cybenko.
\newblock Approximation by superpositions of a sigmoidal function.
\newblock {\em Mathematics of Control, Signals, and Systems}, 2(4):303--314,
  1989.

\bibitem{hornik1989multilayer}
Kurt Hornik, Maxwell~B Stinchcombe, and Halbert White.
\newblock Multilayer feedforward networks are universal approximators.
\newblock {\em Neural Networks}, 2(5):359--366, 1989.

\bibitem{rezende2015variational}
Danilo~Jimenez Rezende and Shakir Mohamed.
\newblock Variational inference with normalizing flows.
\newblock In {\em ICML}, 2015.

\bibitem{DBLP:conf/aaai/LiptonLG00D18}
Zachary~C. Lipton, Xiujun Li, Jianfeng Gao, Lihong Li, Faisal Ahmed, and
  Li~Deng.
\newblock Bbq-networks: Efficient exploration in deep reinforcement learning
  for task-oriented dialogue systems.
\newblock In {\em AAAI}, pages 5237--5244, 2018.

\bibitem{tang2018exploration}
Yunhao Tang and Shipra Agrawal.
\newblock Exploration by distributional reinforcement learning.
\newblock In {\em IJCAI}, 2018.

\bibitem{moerland2018the}
Thomas~M Moerland, Joost Broekens, and Catholijn~M Jonker.
\newblock The potential of the return distribution for exploration in {RL}.
\newblock {\em arXiv: Learning}, 2018.

\bibitem{mavrin2019distributional}
Borislav Mavrin, Shangtong Zhang, Hengshuai Yao, Linglong Kong, Kaiwen Wu, and
  Yaoliang Yu.
\newblock Distributional reinforcement learning for efficient exploration.
\newblock {\em arXiv: Learning}, 2019.

\bibitem{dabney2017distributional}
Will Dabney, Mark Rowland, Marc~G Bellemare, and Remi Munos.
\newblock Distributional reinforcement learning with quantile regression.
\newblock In {\em AAAI}, 2018.

\bibitem{schaul2015prioritized}
Tom Schaul, John Quan, Ioannis Antonoglou, and David Silver.
\newblock Prioritized experience replay.
\newblock In {\em ICLR}, 2016.

\end{thebibliography}

\newpage

\section{Appendix}\label{Appendix}

\subsection{Chain MDP}\label{chainmdp}
\begin{figure}[H]
\centerline{\includegraphics[width=0.7\columnwidth]{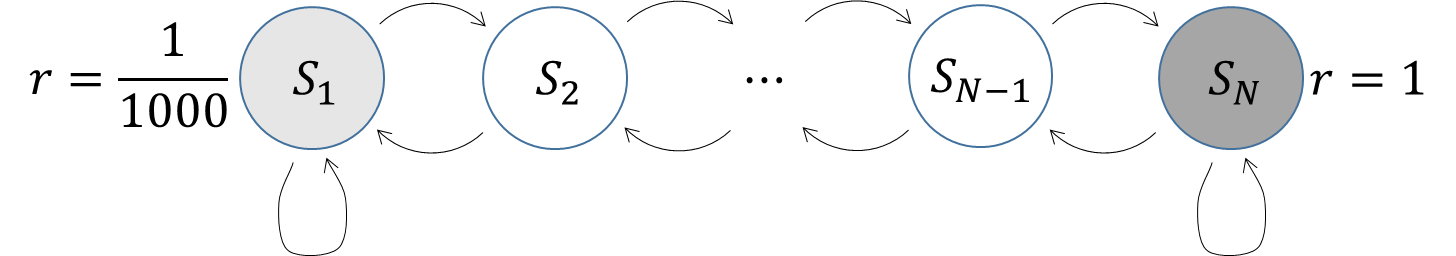}}
\caption{Illustration of a chain MDP.}\label{cmdp}
\end{figure}

As introduced in \cite{osband2016deep}, in chain MDP task (Fig.~\ref{cmdp}), there are $N$ states from $s_1$ to $s_N$. The agent starts from $s_2$ and moves $N+9$ steps choosing left or right in each step. By visiting $s_1$ the agent gets reward $r=\frac{1}{1000}$ or it obtains reward $r=1$ when reaching $s_N$. Obviously, the local optimal policy is repeatedly visiting $s_1$ and the global optimum is consistently choosing right and visit $s_N$ (for $N \le 10991$).

\subsection{Visit Count for Chain MDP}\label{visitcount}

Here we present the state visit counts of DQN, VDQN, NoisyNet, and AVDQN for chain MDP $N=8, 32$ and $128$ in Fig.~\ref{fig7}. We set the visit count $c_n=1$ if state $s_n$ is reached in one episode and $c_n=0$ otherwise for $n=1\dots N$. The average of $c_n$ within 10 episodes is regarded as the approximation of visit probability $p_n$ of state $s_n$.

For $N=8$,  DQN, VDQN, and AVDQN show moderate exploration capability as the probability of visiting $s_N$ increases to nearly 1. DQN and AVDQN are more stable while the performance of VDQN declines after 75s. Meanwhile, NoisyNet achieves worse exploration since its probability of visiting $s_N$ is only around $0.6$.

For $N=32$, DQN occasionally has a nontrivial probability of visiting $s_{\frac{N}{2}}$ because of the $\varepsilon$-greedy random exploration, but it gets lost in halfway and even cannot visit $s_N$. VDQN and NoisyNet can explore deeper than DQN with an around $0.5$ probability of visiting $s_N$. In comparison, AVDQN converges to the global optimum. 

For $N=128$, DQN cannot explore deeply and reach $s_{\frac{N}{2}}$, let alone visiting $s_N$. VDQN and NoisyNet can explore more than DQN, however their performance is unsteady and cannot converge to the global optimum given limited time. In comparison, AVDQN makes progress steadily with an increasing probability of going beyond $s_{\frac{N}{2}}$ and visiting $s_N$ in the pre-train stage, and then converge to near global optimum quickly in the fine-tune stage. Fig.~\ref{fine_tune_MDP128} shows the state visit probability for the chain MDP $N=128$ in the fine-tune stage obtained by applying AVDQN.

\begin{figure}[htbp]
\subfigtopskip=3pt
\subfigbottomskip=3pt
\setlength{\belowcaptionskip}{-10pt}
\subfigure[Chain MDP $N=8$: DQN]
{\begin{minipage}{0.48\textwidth}
\centering
\includegraphics[width =1\columnwidth]{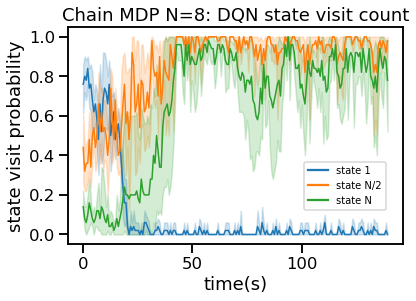}
\end{minipage}}%
\subfigure[Chain MDP $N=8$: VDQN]{
\begin{minipage}{0.48\textwidth}
\includegraphics[width = 1\columnwidth]{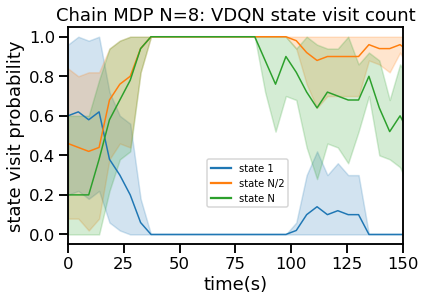}\centering
\end{minipage}}
\end{figure}

\begin{figure}[H]
\subfigure[Chain MDP $N=8$: AVDQN]{
\begin{minipage}{0.48\textwidth}
\includegraphics[width = 1\columnwidth]{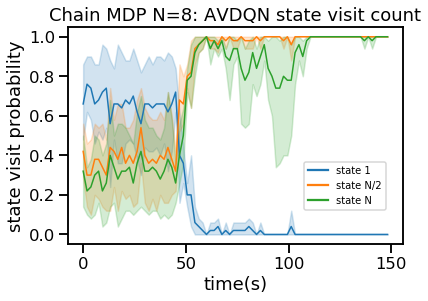}\centering
\end{minipage}}
\subfigure[Chain MDP $N=8$: NoisyNet]{
\begin{minipage}{0.48\textwidth}
\includegraphics[width = 1\columnwidth]{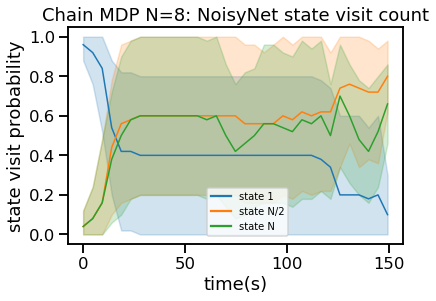}\centering
\end{minipage}}

\subfigure[Chain MDP $N=32$: DQN]
{\begin{minipage}{0.48\textwidth}
\centering
\includegraphics[width =1\columnwidth]{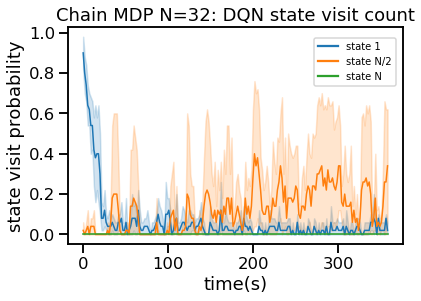}
\end{minipage}}%
\subfigure[Chain MDP $N=32$: VDQN]{
\begin{minipage}{0.48\textwidth}
\includegraphics[width = 1\columnwidth]{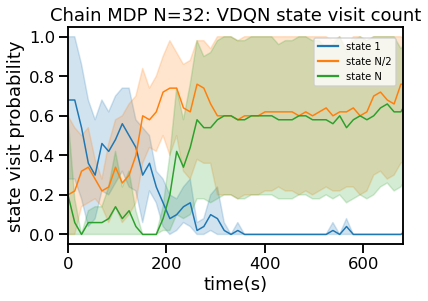}\centering
\end{minipage}}
\subfigure[Chain MDP $N=32$: AVDQN]{
\begin{minipage}{0.48\textwidth}
\includegraphics[width = 1\columnwidth]{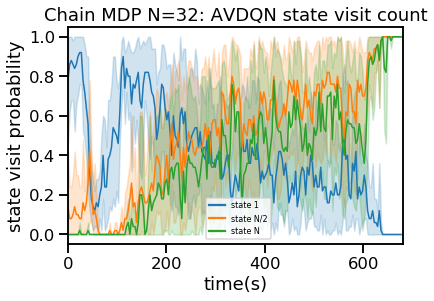}\centering
\end{minipage}}
\subfigure[Chain MDP $N=32$: NoisyNet]{
\begin{minipage}{0.48\textwidth}
\includegraphics[width = 1\columnwidth]{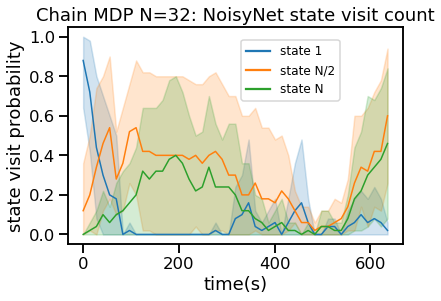}\centering
\end{minipage}}

\subfigtopskip=0pt
\subfigbottomskip=0pt
\subfigure[Chain MDP $N=128$: DQN]
{\begin{minipage}{0.48\textwidth}
\centering
\includegraphics[width =1\columnwidth]{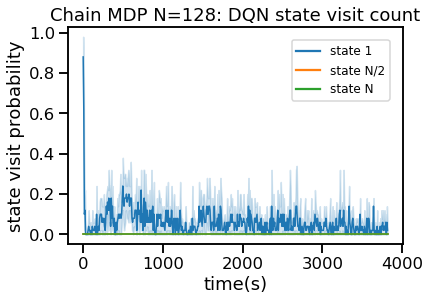}
\end{minipage}}%
\subfigure[Chain MDP $N=128$: VDQN]{
\begin{minipage}{0.48\textwidth}
\includegraphics[width = 1\columnwidth]{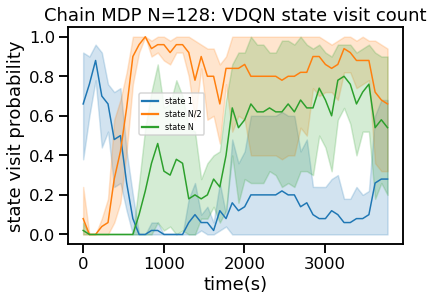}\centering
\end{minipage}}
\end{figure}

\begin{figure}[H]
\subfigure[Chain MDP $N=128$: AVDQN]{
\begin{minipage}{0.48\textwidth}
\includegraphics[width = 1\columnwidth]{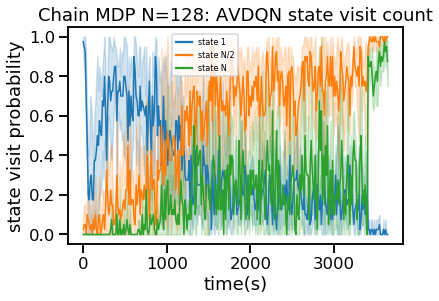}\centering
\end{minipage}}
\subfigure[Chain MDP $N=128$: NoisyNet]{
\begin{minipage}{0.48\textwidth}
\includegraphics[width = 1\columnwidth]{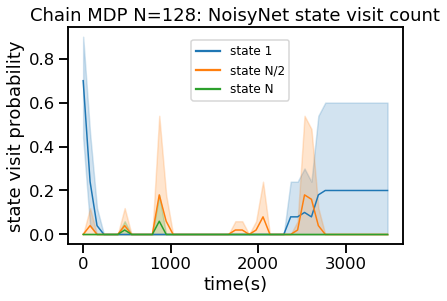}\centering
\end{minipage}}
\caption{The approximate visit probability of state $s_1, s_{\frac{N}{2}}$ and $s_N$ in the chain MDP $N=8$, 32 and 128 averaged over 5 random initializations. Let $c_n=1$ if state $s_n$ is visited in one episode and the average of $c_n$ in 10 episodes is used as approximation of visit probability $p_n$.}\label{fig7}
\end{figure}

\begin{figure}[H]
\centering
\includegraphics[scale=0.45]{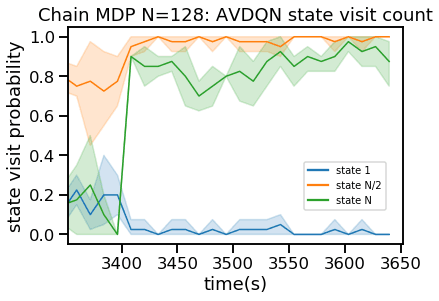}
\caption{The approximate visit probability of $s_1, s_{\frac{N}{2}}$ and $s_N$ in the chain MDP $N=128$ in the fine-tune stage.}\label{fine_tune_MDP128}
\end{figure}


\end{document}